\documentclass[pdflatex,sn-vancouver-ay]{sn-jnl}
\usepackage{graphicx}
\usepackage{multirow}
\usepackage{xcolor}
\usepackage{textcomp}
\usepackage{amsmath}
\usepackage{longtable}
\usepackage{geometry}
\usepackage{array}

\title{\textbf{Limits of trust in medical AI}}

\author{\fnm{Joshua} \sur{Hatherley}}\email{jjh@hum.ku.dk}

\affil{\orgdiv{Center for the Philosophy of AI}, \orgname{University of Copenhagen, Denmark}}

\date{March 27th, 2020}

\abstract{Artificial intelligence (AI) is expected to revolutionise the practice of medicine. Recent advancements in the field of deep learning have demonstrated success in variety of clinical tasks: detecting diabetic retinopathy from images, predicting hospital readmissions, aiding in the discovery of new drugs, etc. AI’s progress in medicine, however, has led to concerns regarding the potential effects of this technology upon relationships of trust in clinical practice. In this paper, I will argue that there is merit to these concerns, since AI systems can be relied upon, and are capable of reliability, but cannot be trusted, and are not capable of trustworthiness. Insofar as patients are required to rely upon AI systems for their medical decision-making, there is potential for this to produce a deficit of trust in relationships in clinical practice.

\bigskip

\bigskip

This is a pre-print of: Hatherley, Joshua. 2020. Limits of trust in medical AI. \textit{Journal of Medical Ethics} 46(7): 478--481. \href{https://doi.org/10.1136/medethics-2019-105935}{10.1136/medethics-2019-105935}}

\begin{document}

\maketitle

\section{Introduction}

Artificial intelligence (AI) is expected to revolutionise the practice of medicine. Recent advancements in the field of deep learning have demonstrated success in variety of clinical tasks; for instance, detecting diabetic retinopathy from images \citep{gulshan2016development}, predicting hospital readmissions \citep{caruana2015intelligible}, and aiding in the discovery of new drugs \citep{ramsundar2015massively}. It has been suggested that AI will facilitate a variety of improvements in medical practice, ranging from economic savings to the improvement of empathetic communication between doctors and patients, from increased productivity to greater professional satisfaction, and from improved health outcomes to an amplified rate of discovery in medical research \citep{topol2019deep}. AI’s progress in medicine, however, has led to concerns regarding the potential effects of this technology upon relationships of trust, particularly between doctors and patients \citep{nundy2019promoting}. In this paper, I will argue that there is merit to these concerns, since AI systems are not the appropriate objects of trust under any familiar philosophical accounts of trust. This is critical, since, as I will argue in section \ref{4}, AI systems are likely to displace the epistemic authority of human clinicians if they come to exceed them in performance. As such, I will argue that insofar as patients are required to rely upon AI systems for their medical decision-making, AI threatens to produce a deficit in trusting clinical relationships between doctors and patients.

\section{Trust in medicine}

Trust has both intrinsic and instrumental significance in medicine.\footnote{Two kinds of trust are discussed in relation to medicine and clinical practice: interpersonal and social \citep{mechanic1996changing}. Interpersonal trust concerns trust between persons (e.g. between doctors and patients), while social trust is more general and abstract, directed towards groups and institutions as opposed to individuals (e.g. between patients and a particular hospital or the medical institution more generally). In this paper, I leave the issue of social trust to one side in order to focus in on medical AI and interpersonal trust. All references to trust will henceforth refer exclusively to interpersonal trust.}  Intrinsically, trust is what imbues the doctor-patient relationship with its uniqueness and importance. A patient comes to a physician in a state of sickness and vulnerability, and is thereby forced to place their trust in another person to treat them with competence and, ideally, empathy and care. This vulnerability of the patient is what imbues the relationship with inherent value, since “trust is inseparable from vulnerability, in that there is no need for trust in the absence of vulnerability” \citep[615]{hall2001trust}. The vulnerability of the patient, and the resulting power of the physician, imbue the physician with a fiduciary obligation to behave in a morally upright and appropriate manner, to use their authority in the service of the patient as opposed to themselves or some other end.

In contrast, trust also has instrumental value in medicine. Firstly, because patients are more likely to accept and behave in accordance with their physician’s judgement if they have a trusting relationship with them. They are more likely to demonstrate “willingness to seek care, reveal sensitive information, submit to treatment, participate in research, adhere to treatment regimens, remain with a physician, and recommend physicians to others” \citep[614]{hall2001trust}. Secondly, it is speculated that trusting doctor-patient relationships have a number of therapeutically valuable effects upon patients – improved patient outcomes and placebo effects, for example. Finally, a good physician is one that can demonstrate care for their patients, and patients are more likely to feel that they have been adequately cared for when they trust the person caring for them.

\section{AI in medicine}

AI’s effect on relations of trust between doctors and patients is bound up with the precise role that AI may come to occupy in medical practice and the epistemic authority that it comes to hold in clinical decision-making procedures. If AI systems are eventually adopted as merely another tool at the clinician’s disposal – akin to a stethoscope, thermometer, or blood pressure monitor – the effect of these systems on trust would likely be minimal. Patients, of course, would rely on the accuracy of these tools, but their trust would be staked in the judgement of the human physician who is interprets their outputs and incorporates them into their own clinical judgements. However, recent developments in areas such as deep learning suggest that the epistemic authority of human clinicians in clinical decision-making will be challenged by the use of AI in medicine.

Researchers in AI are working busily to develop AI systems that can surpass the performance of human clinicians in diagnosis, prognosis, and treatment selection \citep{topol2019deep} – three of the four fundamental tasks of the clinician, according to Eric Cassell \citeyearpar{cassell2002doctoring}.\footnote{The fourth is the identification of causes. Given that AI systems based on neural networks learn from correlations alone, their capacity to illuminate underlying causes of illness is limited \citep{marcus2018deep}.}  Indeed, a recent systematic review and meta-analysis comparing the performance of deep learning AI systems to human clinicians found that deep learning AI systems already match the accuracy of human clinicians in the performance of certain diagnostic tasks \citep{liu2019comparison}.\footnote{Importantly, the study identified a number of troubling methodological limitations in the broader literature comparing the performance of human clinicians to deep learning AI systems, so this finding ought to be taken with a grain of salt. Most alarmingly, of the 31,587 scholarly articles returned on a search for articles comparing the performance of deep learning systems and human clinicians, only 14 compared performance between the two groups on the same test dataset.}   If AI succeeds in surpassing the performance of human clinicians in such principal medical tasks, how might this effect the epistemic authority of human clinicians in clinical practice?

The prospect gestures at an important problem currently faced in the sciences, which Paul Humphreys has called our ‘anthropocentric predicament’. Humphreys argues that advanced technologies have produced a situation in which 

\begin{quote}
    an exclusively anthropocentric epistemology is longer appropriate because there now exist superior, non-human, epistemic authorities. So we are now faced with a problem, which we can call the anthropocentric predicament, of how we, as humans, can understand and evaluate computationally based scientific methods that transcend our own abilities” \citep[617]{humphreys2009philosophical}.
\end{quote}

There have been two principal kinds of response to medicine’s anthropocentric predicament in the wake of medical AI, which I will refer to as substitionism and extensionism. Substitutionists argue that advanced AI will eventually make doctors obsolete by surpassing them in the performance of key clinical tasks and roles \citep{goldhahn2018could}. Extensionists, in contrast, argue that AI will simply extend and improve upon the capabilities and competencies of human clinicians without replacing them outright. In particular, this is because AI systems lack emotional intelligence and empathy, abilities that are essential in the delivery of healthcare, meaning that a human presence will still be essential \citep{krittanawong2018rise}. Yet among both camps, the likely disruptive impact – what Liu and colleagues have labelled a “seismic shift” \citep[115]{liu2018time} – that AI will have medicine is widely undisputed. Although extensionists rally against the substitution of clinicians, the likelihood of their displacement in key clinical roles is often acknowledged. For instance, Eric Topol, a principal physician advocate for the use of AI in medicine, along with his colleague Saurabh Jha, claim that “[j]obs are not lost; rather, roles are redefined; humans are displaced to tasks needing a human element” \citep[2354]{jha2016adapting}.

This displacement of the roles of human clinicians in the wake of advanced medical AI reflects a displacement of their epistemic authority. Indeed, if AI surpasses the performance of clinicians in key clinical tasks, doctors will have an epistemic obligation to defer to the judgements of the machine or align their judgements with the AI in their clinical decision-making \citep{grote2020ethics}. As Bjerring and Busch have argued, “if a practitioner knows of an epistemic source that is more knowledgeable, more accurate, and more reliable in decision-making, she should treat it as an expert and align her verdicts with those of the source” \citep[351]{bjerring2021artificial}. This displacement of the epistemic authority of clinicians would be necessary to realise some of the goals of the introduction of AI in medicine. Aside from the possible reduction of burdensome administrative tasks and the improvement of cost-efficiency in medicine, a primary motivation for research into medical AI is the potential to reduce the alarming prevalence of wastefulness and human error in medical practice \citep{topol2019deep, makary2016medical}. In order to achieve this, it would be necessary in most instances for human clinicians to give more weight to the outputs of a supremely reliable AI system over their own clinical intuitions and judgements. 

The displacement of clinicians from a position of epistemic authority in clinical decision-making has important implications for relations of trust between patients and doctors, since it implies a displacement of patient trust from human clinicians to AI systems. In the next section, I will argue that this displacement of trust from humans to machines could lead to shallow relations of trust in clinical practice that are lacking in important respects.

\section{Trust in AI}\label{4}

Trust has been a central topic of concern in the debate over AI and its many applications, with some private corporations and research organisations releasing guidelines for the development of trust and trustworthiness in AI \citep{high20219ethics, ibm2018principles}. Concerns over the ‘black box’ nature of some AI systems – particular deep learning – along with the threat of algorithmic bias have pushed the issue of trust to the forefront of debate \citep{sparrow2020high}. But what does it mean to say that one trusts an AI, or that an AI is trustworthy? A key response to this question has been to emphasise the centrality of reliability in trust. Alex John London claims that “[i]f the goal is to secure trust among stakeholders, then the accuracy of a system relative to viable alternatives must be a central concern” \citep[18]{london2019artificial}. Similarly, Zachary Lipton claims that if trust is “simply confidence that a model will perform well [… then] a sufficiently accurate model should be demonstrably trustworthy” \citep[7]{lipton2018mythos}.

But is confidence in someone or something’s accuracy or reliability sufficient for trust? According to many accounts of interpersonal trust have been proposed in the philosophical literature, the answer to this question is no. According to these accounts, trusting someone to do \textit{x} is more than merely relying on them to do \textit{x}. Consider the following two scenarios:

\begin{enumerate}
    \item Stan, a thief, is planning a burglary. He has observed a wealthy homeowner, Jane, leaving her home at 9am and returning at 7pm every Monday for the past month. Stan is hoping to go through with his planned burglary next Monday, and is relying upon Jane to continue her pattern in order for his burglary to be successful.
    \item Brendan has a chronic illness that causes him significant pain and suffering. His illness is managed by his regular GP, Dr. Smith. Dr. Smith has supported Brendan through his illness for 15 years. Brendan has recently been experiencing significantly more pain than usual, which is causing him extreme discomfort. He makes an appointment with Dr. Smith, confident that she will be able to help him relieve this pain in some way. 
\end{enumerate}

In the scenario (1), although the thief relies upon Jane to leave her house at 9am, it seems inappropriate to say that the thief trusts Jane to do so in the same way that Brendan trusts Dr. Smith to successfully treat his illness in scenario (2), despite the fact that Brendan also relies upon Dr. Smith. How do we explain this intuition? What makes trusting someone more than merely relying upon them? 

Russell Hardin \citeyearpar{hardin2002trust} argues that reliance is insufficient for trust because trusting someone also requires a belief that one’s interests are encapsulated in the interests of the trusted person. “What matters”, claims Hardin, “[..] is not merely my expectation that you will act in certain ways but also my belief that you have the relevant motivations to act in those ways, that you deliberately take my interests into account because they are mine” \citep[11]{hardin2002trust}. For Hardin, trust requires not only a predictive expectation on the part of the truster, but also a belief that one’s interests are encapsulated in the interests of the trusted person and that the trusted person has the right motivations for action. Indeed, Hardin claims that “I would not, in our usual sense, trust a fully programmed automaton, even if it were programmed to discover and attempt to serve my interests – although I might come to rely heavily on it” \citep[12]{hardin2002trust}. 

Having the right kind of motivations for action is an important part of many other influential accounts of trust. Annette Baier, for instance, argues that reliance underdetermines trust because trust “seems to be reliance on [the trusted person’s] good will toward one, as distinct from their dependable habits, or only on their dependably exhibited fear, anger, or other motives compatible with ill will toward one, or on motives not directed to one at all” \citep[234]{baier1986trust}. This emphasis upon the good will of the trusted person is also central to Karen Jones’ account, wherein she claims that “to trust someone is to have an attitude of optimism about her goodwill and to have the confident expectation that, when the need arises, the one trusted will be directly and favorably moved by the thought that you are counting on her” \citep[5-6]{jones1996trust}. If the right kind of motivations are necessary for the kind of trust that we would usually recognise as interpersonal trust, then AI systems would not appear to be the appropriate objects of this kind of trust. Unlike a human clinician, AI systems have no goodwill towards us, nor any motivation to act in our interests. This may be at least part of the reason that some people may be uncomfortable with the idea of placing their trust in an AI for important medical decisions or tasks. 

Additionally, other philosophical accounts of trust distinguish between trust and reliance on the basis of normative and descriptive expectations: I rely on you when I predict that you will behave in a certain way, though I trust you when I judge that you ought to behave in a certain way \citep{nickel2012risk}. Trusting someone, that is, generates an obligation on behalf of the trusted person to (at least genuinely attempt to) do what one is trusting them to do. There are some important limitations to this claim, e.g. in circumstances where the trust that one has in another is misguided or unwelcome. Suppose, for instance, that one were to place their trust in a friend who is a dermatologist to remove their wisdom teeth. Trusting the dermatologist for this procedure would appear quite mistaken, given that the dermatologist does not have the expertise or competency to perform this task. Nor, presumably, would the dermatologist welcome this trust in any way. 

But outside of this and other somewhat fanciful scenarios, clinicians do in fact have an obligation to perform those tasks that have been entrusted to them, providing of course that this trust has been communicated to them. This is precisely the nature of fiduciary obligations in medicine. If this is true, another limitation of trusting AI would also be demonstrated, since AI systems are not the appropriate objects of moral responsibility. In order for an agent to be morally responsible for an action, they must be blameworthy when they fail to come through on that action. But if an AI system were to incorrectly diagnose a patient, leading to their avoidable death, it would appear misguided or inappropriate to blame the AI for its error. Rather, one would generally look to the designers, the supervising clinician, the hospital, etc. in order to apportion blame. Trusting a clinician generates a moral responsibility on behalf of the clinician, while trusting an AI system generates a moral responsibility on behalf of seemingly anyone but the AI system.

These considerations highlight two important deficits in relations between patients and medical AI systems that each stem from a lack of agency on the part of the AI. Firstly, AI systems lack the right kind of motivation for trust - either in the form of encapsulated interest or a sense of good will – since they lack motivation entirely. Secondly, relations with AI systems cannot be said to be trusting relations, as one might have with a human clinician, since trust generates normative obligations that cannot be borne by an AI. To say that one can trust an AI is thus akin to saying that one can trust a naturally occurring phenomenon. Although I am supremely confident that tomorrow the sun will rise in the east and set in the west, there is not familiar sense in which I could reasonably said to trust the sun to do so. Trusting relations, in other words, are exclusive to beings with agency, meaning that the displacement of human clinicians from a position of epistemic authority and privilege in the clinical encounter threatens to lead to relations of trust that are shallow or deficient in important respects within medical practice.

\section{Conclusion}

To say that one can trust an AI system, or that the AI is trustworthy, is merely to say that one can rely on the AI system, or that the system is reliable. Yet as we have seen, reliability is insufficient to generate a relation of trust under any of its familiar philosophical notions, which all require characteristics essential and exclusive to beings with a form of agency. What does this mean for the pursuit of ‘Trustworthy AI’ initiated by the European Union’s High-Level Expert Group on AI (HLEG AI) \citeyearpar{high20219ethics}? Although valuable, the pursuit of trustworthy AI represents a notable conceptual misunderstanding, since AI systems are not the appropriate objects of trust or trustworthiness. Interestingly, this has also been suggested by a key member of the HLEG AI, Thomas Metzinger \citeyearpar{metzinger2019eu}. Rather than trustworthy AI, this pursuit may be better served by being reframed in terms of reliable AI, reserving the label of ‘trust’ for reciprocal relations between beings with agency.

In contrast to AI, therefore, human clinicians can offer their patients the kind of rich interpersonal trust that imbues the doctor-patient relationship with its uniqueness and significance. Insofar as patients come to rely upon AI systems for important medical assessments and decisions as opposed to human clinicians, they may be sacrificing opportunities for trusting relationships in medicine. A more thoughtful engagement is needed with the potential effects of AI on medical practice to further understand the implications of this technology, so that it can be deployed is such a way as to reap its potential benefits whilst retaining those aspects of medicine – such as trust – that are particularly valuable for its functioning.

\section*{Acknowledgments}
Thanks to Rob Sparrow for comments on an earlier draft of this paper. Research for this paper was funded through Australian Government Research Training Program Scholarship.

\bibliography{main}

\begin{thebibliography}{28}
\providecommand{\natexlab}[1]{#1}
\providecommand{\doi}[1]{\url{https://doi.org/#1}}
\providecommand{\url}[1]{\texttt{#1}}
\providecommand{\urlprefix}{}

\bibitem[{Baier(1986)Baier, Annette}]{baier1986trust}
Baier A.
\newblock Trust and antitrust.
\newblock Ethics. 1986;96(2):231--260.

\bibitem[{Bjerring and Busch(2021)Bjerring, Jens Christian and Busch, Jacob}]{bjerring2021artificial}
Bjerring JC, Busch J.
\newblock Artificial intelligence and patient-centered decision-making.
\newblock Philosophy \& Technology. 2021;34:349--371.

\bibitem[{Caruana et~al.(2015)Caruana, Rich and Lou, Yin and Gehrke, Johannes and Koch, Paul and Sturm, Marc and Elhadad, Noemie}]{caruana2015intelligible}
Caruana R, Lou Y, Gehrke J, Koch P, Sturm M, Elhadad N.
\newblock Intelligible models for healthcare: Predicting pneumonia risk and hospital 30-day readmission.
\newblock In: Proceedings of the 21th ACM SIGKDD International Conference on Knowledge Discovery and Data Mining; 2015. p. 1721--1730.

\bibitem[{Cassell(2004)Cassell, Eric J}]{cassell2002doctoring}
Cassell EJ.
\newblock The nature of suffering: And the goals of medicine.
\newblock Oxford University Press; 2004.

\bibitem[{Goldhahn et~al.(2018)Goldhahn, J{\"o}rg and Rampton, Vanessa and Spinas, Giatgen A}]{goldhahn2018could}
Goldhahn J, Rampton V, Spinas GA.
\newblock Could artificial intelligence make doctors obsolete?
\newblock BMJ. 2018;363:k4563.

\bibitem[{Grote and Berens(2020)Grote, Thomas and Berens, Philipp}]{grote2020ethics}
Grote T, Berens P.
\newblock On the ethics of algorithmic decision-making in healthcare.
\newblock Journal of Medical Ethics. 2020;46(3):205--211.

\bibitem[{Gulshan et~al.(2016)Gulshan, Varun and Peng, Lily and Coram, Marc and Stumpe, Martin C and Wu, Derek and Narayanaswamy, Arunachalam and Venugopalan, Subhashini and Widner, Kasumi and Madams, Tom and Cuadros, Jorge and others}]{gulshan2016development}
Gulshan V, Peng L, Coram M, Stumpe MC, Wu D, Narayanaswamy A, et~al.
\newblock Development and validation of a deep learning algorithm for detection of diabetic retinopathy in retinal fundus photographs.
\newblock JAMA. 2016;316(22):2402--2410.

\bibitem[{Hall et~al.(2001)Hall, Mark A and Dugan, Elizabeth and Zheng, Beiyao and Mishra, Aneil K}]{hall2001trust}
Hall MA, Dugan E, Zheng B, Mishra AK.
\newblock Trust in physicians and medical institutions: What is it, can it be measured, and does it matter?
\newblock The Milbank Quarterly. 2001;79(4):613--639.

\bibitem[{Hardin(2002)Hardin, Russell}]{hardin2002trust}
Hardin R.
\newblock Trust and trustworthiness.
\newblock Russell Sage Foundation; 2002.

\bibitem[{{High-Level Expert Group on Artificial Intelligence}(2019)}]{high20219ethics}
{High-Level Expert Group on Artificial Intelligence}.
\newblock Ethics guidelines for trustworthy {AI}.
\newblock European Commission; 2019.

\bibitem[{Humphreys(2009)Humphreys, Paul}]{humphreys2009philosophical}
Humphreys P.
\newblock The philosophical novelty of computer simulation methods.
\newblock Synthese. 2009;169:615--626.

\bibitem[{IBM(2018)}]{ibm2018principles}
IBM.: {IBM}'s principles for data trust and transparency; 2018.
\newblock \urlprefix\url{https://www.ibm.com/blogs/policy/trust-principles/}.

\bibitem[{Jha and Topol(2016)Jha, Saurabh and Topol, Eric J}]{jha2016adapting}
Jha S, Topol EJ.
\newblock Adapting to artificial intelligence: Radiologists and pathologists as information specialists.
\newblock JAMA. 2016;316(22):2353--2354.

\bibitem[{Jones(1996)Jones, Karen}]{jones1996trust}
Jones K.
\newblock Trust as an affective attitude.
\newblock Ethics. 1996;107(1):4--25.

\bibitem[{Krittanawong(2018)Krittanawong, Chayakrit}]{krittanawong2018rise}
Krittanawong C.
\newblock The rise of artificial intelligence and the uncertain future for physicians.
\newblock European Journal of Internal Medicine. 2018;48:e13--e14.

\bibitem[{Lipton(2018)Lipton, Zachary C}]{lipton2018mythos}
Lipton ZC.
\newblock The mythos of model interpretability.
\newblock Queue. 2018;16(3):31--57.

\bibitem[{Liu et~al.(2019)Liu, Xiaoxuan and Faes, Livia and Kale, Aditya U and Wagner, Siegfried K and Fu, Dun Jack and Bruynseels, Alice and Mahendiran, Thushika and Moraes, Gabriella and Shamdas, Mohith and Kern, Christoph and others}]{liu2019comparison}
Liu X, Faes L, Kale AU, Wagner SK, Fu DJ, Bruynseels A, et~al.
\newblock A comparison of deep learning performance against health-care professionals in detecting diseases from medical imaging: A systematic review and meta-analysis.
\newblock The Lancet Digital Health. 2019;1(6):e271--e297.

\bibitem[{Liu et~al.(2018)Liu, Xiaoxuan and Keane, Pearse A and Denniston, Alastair K}]{liu2018time}
Liu X, Keane PA, Denniston AK.
\newblock Time to regenerate: The doctor in the age of artificial intelligence.
\newblock Journal of the Royal Society of Medicine. 2018;111(4):113--116.

\bibitem[{London(2019)London, Alex John}]{london2019artificial}
London AJ.
\newblock Artificial intelligence and black-box medical decisions: Accuracy versus explainability.
\newblock Hastings Center Report. 2019;49(1):15--21.

\bibitem[{Makary and Daniel(2016)Makary, Martin A and Daniel, Michael}]{makary2016medical}
Makary MA, Daniel M.
\newblock Medical error -- {T}he third leading cause of death in the {US}.
\newblock BMJ. 2016;353:i2139.

\bibitem[{Marcus(2018)Marcus, Gary}]{marcus2018deep}
Marcus G.
\newblock Deep learning: A critical appraisal.
\newblock arXiv preprint. 2018;1801.00631:1--27.

\bibitem[{Mechanic(1996)Mechanic, David}]{mechanic1996changing}
Mechanic D.
\newblock Changing medical organization and the erosion of trust.
\newblock The Milbank Quarterly. 1996;74(2):171--189.

\bibitem[{Metzinger(2019)Metzinger, Thomas}]{metzinger2019eu}
Metzinger T.: {EU} guidelines: Ethics washing made in Europe; 2019.
\newblock \urlprefix\url{https://www.tagesspiegel.de/politik/eu-guidelines-ethics-washing-made-in-europe/24195496.html}.

\bibitem[{Nickel and Vaesen(2012)Nickel, Philip J and Vaesen, Krist}]{nickel2012risk}
Nickel PJ, Vaesen K.
\newblock In: Risk and trust Springer Dordrecht; 2012. p. 858--876.

\bibitem[{Nundy et~al.(2019)Nundy, Shantanu and Montgomery, Tara and Wachter, Robert M}]{nundy2019promoting}
Nundy S, Montgomery T, Wachter RM.
\newblock Promoting trust between patients and physicians in the era of artificial intelligence.
\newblock JAMA. 2019;322(6):497--498.

\bibitem[{Ramsundar et~al.(2015)Ramsundar, Bharath and Kearnes, Steven and Riley, Patrick and Webster, Dale and Konerding, David and Pande, Vijay}]{ramsundar2015massively}
Ramsundar B, Kearnes S, Riley P, Webster D, Konerding D, Pande V.
\newblock Massively multitask networks for drug discovery.
\newblock arXiv preprint. 2015;1502.02072:1--27.

\bibitem[{Sparrow and Hatherley(2020)Sparrow, Robert and Hatherley, Joshua}]{sparrow2020high}
Sparrow R, Hatherley J.
\newblock High hopes for “{D}eep {M}edicine”? {AI}, economics, and the future of care.
\newblock Hastings Center Report. 2020;50(1):14--17.

\bibitem[{Topol(2019)Topol, Eric}]{topol2019deep}
Topol E.
\newblock Deep medicine: How artificial intelligence can make healthcare human again.
\newblock Hachette UK; 2019.

\end{thebibliography}
\end{document}